\documentclass{article}
\usepackage{ijcai23sty}

\usepackage{times}
\usepackage{soul}
\usepackage{url}
\usepackage[utf8]{inputenc}
\usepackage[small]{caption}
\usepackage{graphicx}
\usepackage{amsmath}
\usepackage{amsthm}
\usepackage{booktabs}
\usepackage{algorithm}
\usepackage{algorithmic}
\usepackage[switch]{lineno}

\usepackage[pagebackref=true,breaklinks=true,colorlinks,bookmarks=false]{hyperref}
\usepackage{xspace}
\usepackage{epsfig}
\usepackage{amssymb}
\usepackage{enumerate}
\usepackage{subcaption}
\usepackage{threeparttable}
\usepackage{multirow,booktabs,makecell}
\newcommand{\eg}{{\emph{e.g.}},\xspace}

\newcommand{\backwardwarp}{\overleftarrow{\mathcal{W}}}

\newcommand*\samethanks[1][\value{footnote}]{\footnotemark[#1]}

\title{Collaborative Neural Rendering Using Anime Character Sheets}

\author{
Zuzeng Lin$^{1,2,}$\thanks{These three authors contribute equally.}
\and
Ailin Huang$^{2,3,}$\samethanks \and
Zhewei Huang$^{2,}$\samethanks
\affiliations
$^{1}$Tianjin University~~~~
$^{2}$Megvii Technology~~~~
$^{3}$Wuhan University~~~~
\emails
{\tt\small transpchan@gmail.com, \{huangailin, huangzhewei\}@megvii.com}\\~\\
\url{https://github.com/megvii-research/IJCAI2023-CoNR}
}
\begin{document}

\maketitle

\begin{abstract}
    Drawing images of characters with desired poses is an essential but laborious task in anime production. Assisting artists to create is a research hotspot in recent years. In this paper, we present the Collaborative Neural Rendering~(CoNR) method, which creates new images for specified poses from a few reference images (AKA Character Sheets). In general, the diverse hairstyles and garments of anime characters defies the employment of universal body models like SMPL~\cite{loper_smpl_2015}, which fits in most nude human shapes. To overcome this, CoNR uses a compact and easy-to-obtain landmark encoding to avoid creating a unified UV mapping in the pipeline. In addition, the performance of CoNR can be significantly improved when referring to multiple reference images, thanks to feature space cross-view warping in a carefully designed neural network. Moreover, we have collected a character sheet dataset containing over $700,000$ hand-drawn and synthesized images of diverse poses to facilitate research in this area. 
\end{abstract}

\section{Introduction}
\begin{figure*}
\centering
\includegraphics[width=1.0\linewidth]{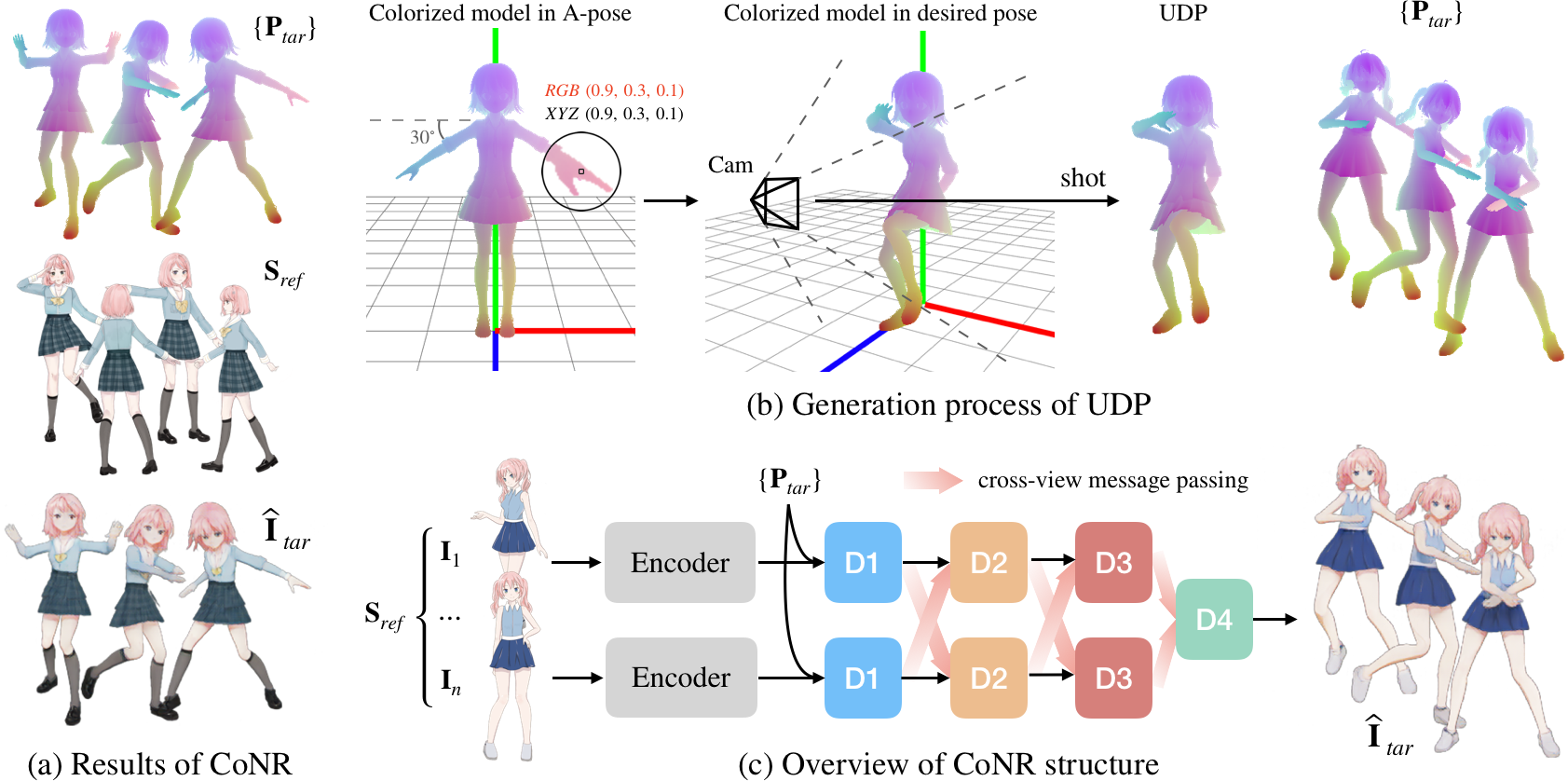}
\caption{\textbf{(a) The results of CoNR.} Based on the desired poses $\{ \textbf{P}_{tar}\}$ and the character sheet $S_{ref}$, CoNR renders new anime images $\hat{\mathbf{I}}_{tar}$. \textbf{(b) The generation process of UDP.} We use the $XYZ$ coordinates of a point on the surface of a 3D model as the $RGB$ value of the point and then color a 3D model. Then, we take a 2D view of the 3D model as UDP. \textbf{(c) Inference pipeline of CoNR.} Reference images $I_1 \cdots I_n\in \mathbf{S}_{ref}$ from the input character sheet are fed into CoNR using modified U-Nets~\protect\cite{ronneberger_u-net_2015} as sub-networks. UDP $\mathbf{P}_{tar}$ is resized and concatenated into each scale of the encoder outputs in all sub-networks. Blocks with the same color share weights. ``D1 to D4" refers to four blocks of the decoder. Each block will receive the averaged message from corresponding blocks in all other sub-networks.
}
\label{fig:acs}
\end{figure*}

2D Animation is one of the essential carriers of art reflecting human creativity. Artists commonly use character sheets to show their virtual character design. A character sheet is the image collection of a specific character with multiple postures observed from different views, as shown in Figure~\ref{fig:acs}. It covers all the appearance details and is widely used to assist the creation of animations or their derived media. Moreover, character sheets allow many artists to cooperate while maintaining the consistency of the design of this character.

Drawing a sequence of anime frames is extremely time-consuming, requiring imagination and expertise. Due to the semantic gap between the character sheet and the desired poses, it is challenging to design a pipeline to draw character images automatically. Some non-photorealistic rendering~(NPR) methods~\cite{gooch2001non} can simulate the artistic style of hand-drawn animation. For example, Toon Shading is widely used in games and animation production. However, it currently still requires a very complex manual design to approximate a specific artistic style. Artists need to manually retouch the shadows, which is too complicated for animators and painters. Therefore, we try to explore a new method of generating pictures in 2D animation.

We formulate the task of rendering a particular character in the desired pose from the character sheet. Based on this formulation, we develop a~\textbf{Collaborative Neural Rendering (CoNR)} model based on convolutional neural network~(CNN). CoNR fully exploits the information available in a provided set of reference images by using feature space cross-view dense correspondence. In addition, CoNR uses the Ultra-Dense Pose~(UDP), an easy-to-construct compact landmark encoding tailored for anime characters. CoNR will not require a unified UV texture mapping~\cite{yoon_pose-guided_2020,gao_portrait_2020} in the pipeline, which may not be done in a consistent method for anime characters. UDP can represent the fine details of characters, such as accessories, hairstyles, or clothing, so as to allow better artistic control. It can also be easily generated with existing 3D computer graphics pipelines to adapt interactive applications. 
Moreover, we collect a character sheet dataset containing over $700,000$ hand-drawn and 3D-synthesized images of diverse poses and appearances. Training on this dataset, CoNR achieves impressive results both on hand-drawn and synthesized images. CoNR can help generate the character in the given pose. Creation of CoNR faithfully based on the character sheets, and it can quickly provide reference of the target pose for the artist. We also introduce how CNN-based UDP Detector generates UDPs from hand-drawn images in the Appendix. We provide a demo video in the supplementary material.

To sum up, our main contributions are:
\begin{itemize}
    \item We formulate a new task, rendering 2D anime character images with desired poses using character sheets.
    \item We introduce a UDP representation for anime characters and collect a large character sheet dataset containing diverse poses and appearances. This dataset is made open-sourced.
	\item We explore a multi-view collaborative inference model, CoNR, to assist in producing impressive anime videos given action sequences specified by UDPs.
\end{itemize}

\section{Related Work}
\subsection{Image Generation and Translation for Anime}
Recent years have seen significant advancement in applying deep learning methods to assist the creation of anime. For example,~\citeauthor{gao_automatic_nodate,zheng_learning_2020} propose to apply realistic lighting effects and shadow for 2D anime automatically;~\citeauthor{wang_learning_nodate} propose to transfer photos to anime style;~\citeauthor{siyao_deep_2021,chen2022eisai} propose frame interpolation tailored for animation. There are also attempts to produce vectorized anime images~\cite{su_vectorization_2021}, similar to the step-by-step manual drawing process. The generative modeling of anime faces has achieved very impressive results~\cite{jin_towards_2017,gokaslan_improving_2019,tseng_modeling_2020,li_surrogate_2021,he_eigengan_2021}. The latent semantics of generative models has also been extensively explored~\cite{shen2021closed}. A modified StyleGAN2 model~\cite{tadne2021} can generate full-body anime images, although it still suffers from artifacts because of the high degree of freedom of the human body.

\subsection{Representation of Human Body}

% We analyze the representation of the human body posture in the real world in multiple data modes.

%\paragraph{Sparse Representations}

Stick-figure of skeletons~\cite{chan_everybody_2019}, SMPL vectors~\cite{loper_smpl_2015}, and heat maps of joints~\cite{paf,siarohin_motion_2021} are widely-used representations obtained from motion capture system. 
However, when these sparse representations are used for anime characters~\cite{khungurn_pose_2016}, they face a new series of challenges: noisy manual annotations, unexpected occlusions caused by the wide variety of characters' clothing and body shapes, and ambiguity due to hand drawing.
Furthermore, the aforementioned human pose representations only represent human joints and body shapes. However, anime characters often require flexible artistic control over other body parts, such as the fluttering of hair and skirts. These representations cannot directly drive these parts.
%\paragraph{Dense Representations}

Human parsers or clothing segmenters~\cite{yoon_pose-guided_2020,gafni_single-shot_2020,chou_template-free_2021} are robust to the uncertainty of joint positions. However, the provided semantic masks are not informative enough to represent the pose or even the orientation of a person. DensePose~\cite{guler_densepose_2018} and UV texture mapping~\cite{yao_densebody_2019,yoon_pose-guided_2020,gao_portrait_2020}, greatly enhance the detail of pose representation on the human body or face by adding a universal definition that essentially unwarps the 3D human body surface into a 2D coordinate system. However, due to the great diversity and rich topology of anime characters, it is difficult to find a general labeling method to unwarp them, which makes existing dense representations still not serve as an off-the-shelf solution for anime-related tasks. One motivation for our work is to find a suitable representation to represent the motion of hair and clothes in principle.

There are also some dense representations of the body, such as CSE~\cite{neverova2020continuous}. CSE addresses the task of detecting a continuous surface coding from an image. Similar ``define-by-training" methods are inapplicable in anime creation since CSE can only be used for reenactment currently. 

% Anime girls have trouble finding the precise location visually where they should cut their skirts and flatten this cone-like object in the same way as others.
% Anime boys get stuck as they are unsure how to consistently handle jeans, kilts, and other non-homeomorphic body shapes. 
% Meanwhile, they have no idea of the number of key points they should use for rabbit or cat ears. Due to the diverse body shapes of anime characters, every body region can require more key points than others. Therefore, 

\subsection{Human Appearance Transfer}

Most of works create vivid body motions or talking heads from only one single image~\cite{gafni_single-shot_2020,sarkar_neural_2021,yoon_pose-guided_2020,siarohin_first_nodate,liu2019liquid}. The learned prior of the human body~\cite{loper_smpl_2015}, head~\cite{blanz_morphable_1999}, or real-world clothing shape and textures~\cite{alldieck_tex2shape_2019} enables the model to solve ill-posed problems like imagining and inpainting the back view even if only the frontal view is available. However, anime has long been featuring a flexible character design leading to high diversity in clothing, body shapes, hairstyles, and other appearance features. A model trained on a huge dataset~(\eg, CLIP~\cite{radford2021learning}) might be able to encode some popular anime character designs implicitly. Still, it is generally more challenging to establish priors or styles of the anime character domain than the real human domain.

There are some attempts~\cite{liu_liquid_nodate} to extend the pose transfer task by utilizing SMPL~\cite{loper_smpl_2015}, a realistic 3D mesh of
the naked human body that is based on skinning and blend-shapes, to combine appearance from different views. Using multiple reference images would, in principle, allow the model to follow the original appearance of the given person and better suit the needs of anime production.

Some recent works utilize neural rendering models~\cite{mildenhall_nerf_2020} which are trained using photometric reconstruction loss and camera poses over multiple images of 3D objects and scenes. Due to their ray-marching nature and capability to in-paint in 3D, they are promising methods in modeling real-world 3D data~\cite{peng_animatable_2021}. These methods are not influenced by or depend on prior knowledge other than the object to be modeled. However, they have not yet made much progress in modeling hand-drawn data like anime character sheets, which less follow strict geometric and physical constraints. 
\iffalse
\begin{figure}[tb]
	\centering
        \includegraphics[width=0.95\linewidth]{image/udp.png}
        \includegraphics[width=0.9\linewidth]{image/mainb.png}
	% \begin{minipage}[t]{0.570\linewidth}
	% 	\centering
	% 	\includegraphics[width=1\linewidth]{image/udpa.png}
	% 	\centerline{(a)}
	% \end{minipage}
	% \begin{minipage}[t]{0.290\linewidth}
	% 	\centering
	% 	\includegraphics[width=1\linewidth]{image/udpb.png}
	% 	\centerline{(b)}
	% \end{minipage}
	\caption{\textbf{The UDPs of an anime character (Kurei Kei)}. UDP uses 3D coordinates at same~\textbf{A-pose} as landmarks. When the anime body changes its pose, the landmark at the corresponding body part will remain the same. Compared to existing dense human-pose representations, UDP bypasses the step of unwarping 3D surfaces onto a 2D UV texture mapping, which may not be done in a consistent method for anime characters.}
	\label{fig:udp}
\end{figure}
\fi

\section{Method}
\subsection{Task Formulation}
\label{subsec:task_formulation}
% We explore this task by first observing real artists drawing anime images. 
Our formulation is inspired by the tasks about drawing or painting art~\cite{huang_learning_2019,su_vectorization_2021}. We consider the character sheet $I_n \in {S}_{ref}$ in a whole. 
A target pose ${P}_{tar}$ representation is also required to provide rendering target for the model.
The task can be formulated as mapping ${S}_{ref}$ to target image $\widehat{I}_{tar}$ with the desired target pose ${P}_{tar}$:
\begin{equation}
\widehat{I}_{tar}=\Phi({P}_{tar},{S}_{ref}).
\end{equation}

We notice that complicated poses, motions, or characters may require more references in ${S}_{ref}$ than others, so dynamically sized ${S}_{ref}$ should be allowed.

% To address our task, 
%Usually, multiple images can be fed in arbitrary order into multiple copies of the same classical convolutional neural network~(CNN) to obtain corresponding inference results. 
% In CINN, however, multiple images in a set are defined in a whole as one single input sample.
% Adding feature-averaging on outputs of all corresponding blocks in multiple copies of a convolutional neural network~(CNN), we obtain a network consisting of a dynamical number of sub-networks. The sub-networks share the same weight and are inter-connected by message passing mechanisms. Due to the commutative nature of addition, changing the order of the sub-networks will not affect the inference results.

% When performing a collaborative inference on such a network, $n$ reference images (or views) in a set are fed into $n$ weight-shared sub-networks, respectively. 
% The sub-networks form a fully connected graph, as illustrated in Figure~\ref{fig:overview}, so that each block of a sub-network would share part of its outputs as messages to corresponding successive blocks in all other sub-networks, in addition to forwarding other outputs to its following blocks like in a classical neural network. 
% To further modulate the cross-view message sent after each block, we apply weighted averaging on messages, where the weights are predicted by CNN and normalized by the number of views~(inspired by \cite{zhou2016view}). 

\subsection{Ultra-Dense Pose}
\label{subsec:Unflattened_dense_pose}
A UDP specifies a character's pose by mapping 2D viewport coordinates to feature vectors, which are 3-tuple floats that continuously and consistently encode body surfaces. In this way, a UDP can be represented as a color image ${P}_{tar}\in  \mathbb{R}^{H\times W\times 3}$ with pixels corresponding to landmarks $L_{(x,y)}\in\mathbb{R}^3$. Non-person areas of the UDP image are masked. It allows better compatibility across a broader range of anime body shapes and enables better artistic control over body details like garment motions.

3D meshes are widely used data representations for anime characters in their game adaptations.
Vertex in a mesh usually comprises corresponding texture coordinate $(u,v)$ or a vertex color $(r,g,b)$. Interpolation over the barycentric coordinates allows triangles to form faces filled with color values or pixels looked up from textures coordinates.

Taking a bunch of anime body meshes standing at the center of the world, we ask them to perform the same~\textbf{A-pose}~(a standard pose) to align the joints. 
To construct UDPs, we remove the original texture and overwrite the color $(r,g,b)$ of each vertex with a landmark, which is currently the world coordinate~$(x,y,z)$, as shown in Figure~\ref{fig:acs}(b).
When the anime body changes its pose, the vertex on the mesh may move to a new position in the world coordinate system, but the landmark at the corresponding body part will remain the same color.

To avoid the difficulty of down-sampling and processing meshes, we convert the modified meshes into 2D images, which are compatible with CNNs.
This is done by introducing a camera, culling on occluded faces, and projecting only the faces visible from the camera into an image. The processed UDP representation is a $H\times W\times 4$ shaped image recorded in floating-point numbers ranging from $0$ to $1$. The four channels include the three-dimensional landmark encodings and one-dimensional occupancy for indicating whether the pixel is on the body.

Three properties of UDP could alleviate the difficulties when creating images of 2D animation:

1) UDP is a detailed 3D pose representation since every piece of surface on the anime body could be automatically assigned with a unique encoding in 3D graphics software. 

2) UDP is a compatible pose representation since the anime characters with similar body shapes will also get out-fits that are consistently pseudo-colorized. 

3) UDP can be obtained directly in all existing 3D editors, game engines, and many other up-streams. 

\begin{figure}[tb]
	\centering
	\includegraphics[width=8.4cm]{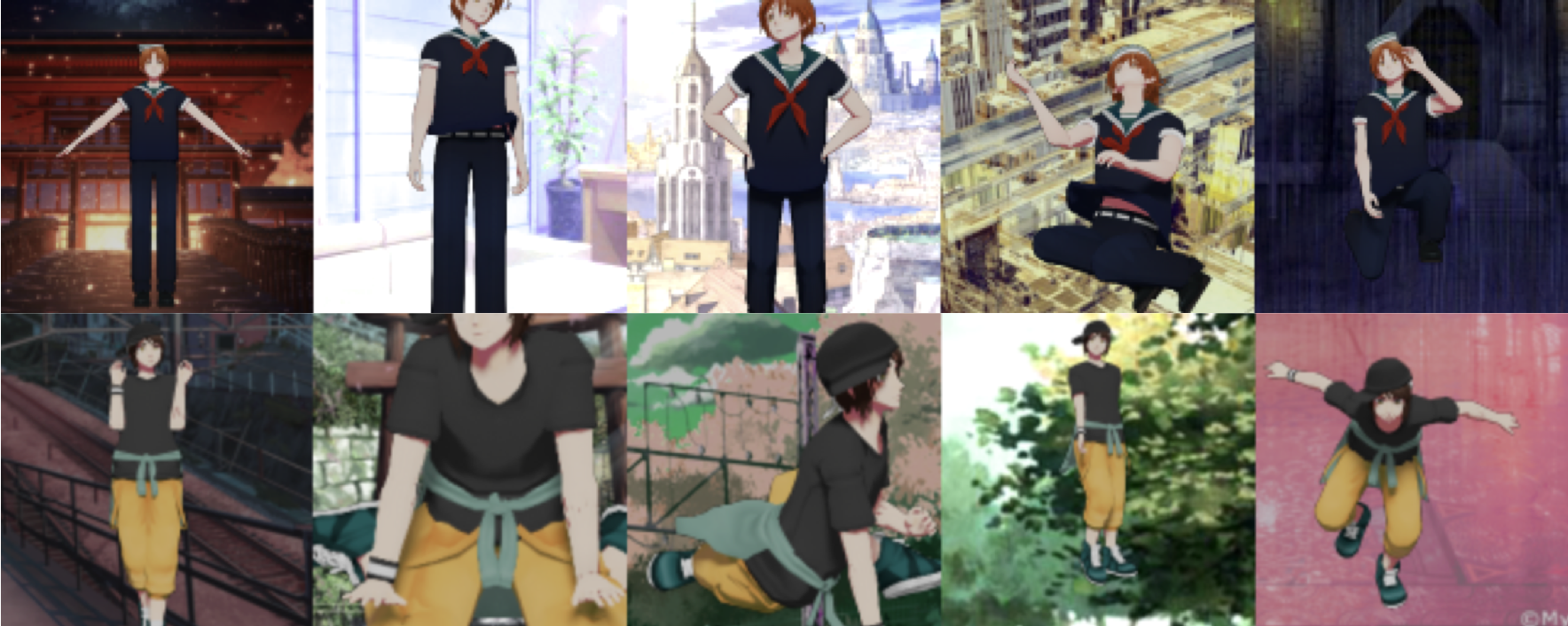}
	\caption{\textbf{Random characters with random backgrounds}.}
	\label{fig:random}
\end{figure}
\subsection{Data Preparation}
\label{subsec:data_preparation}
As character sheets used in the anime-related industries are not yet available to the computer vision community, we built a dataset containing more than $20,000$ hand-drawn anime characters by selecting human-like characters from public datasets~\cite{danbooru2020,pixiv}. We manually perform matting to remove the background from the character with the help of the watershed algorithm~\cite{torralba2010labelme}. We also construct a 2D background dataset containing $4000$ images with a similar method. 
Manually annotating hand-drawn anime images with UDP involves significant hardship. To alleviate the problem of label scarcity, we further construct a synthesized dataset from anime-styled 3D meshes.

Finally, We combine the synthesized dataset with the hand-drawn dataset to obtain both high-quality UDP labels and high diversity of hand-drawn styles. We randomly split the whole dataset by a $1/16$ ratio into the validation and training sets. The split is on a per-anime-character basis, so the validation set contains characters unseen during training. The whole dataset contains over $700,000$ hand-drawn and synthesized images of diverse poses and appearances. We manually exclude content containing excessive nudity that is not suitable for public display. Random characters with a random background are shown in Figure~\ref{fig:random}.

\subsection{Collaborative Neural Rendering}
\label{subsec:implement}

\iffalse
\begin{figure*}[tb]
	\centering
	\includegraphics[width=17.5cm]{image/overview.png}
	\caption{\textbf{Inference pipeline of CoNR.} 
	Reference images $I_1 \cdots I_n\in \mathbf{S}_{ref}$ from the input character sheet are fed into CoNR using modified U-Nets~\cite{ronneberger_u-net_2015} as sub-networks. The same UDP representation $\mathbf{P}_{tar}$ is resized and concatenated into each scale of the encoder outputs in all sub-networks. Blocks with the same color share weights. ``D1 to D4" refers to 4 blocks of the decoder. The sub-networks form a fully connected graph using cross-view message passing. Each block will receive the averaged message from corresponding blocks in all other sub-networks.}\label{fig:overview}
\end{figure*}
\fi

\noindent\textbf{Overview.} We utilize a collaborative inference for a convolutional neural network named \textbf{CINN}, inspired by PointNet~\cite{qi2017pointnet} and Equivariant-SFM~\cite{moran2021deep}. CoNR consists of a CINN renderer and an optional UDP Detector. Figure~\ref{fig:acs}(c) shows the pipeline of the proposed approach. CoNR generates a character image of the desired pose, taking the UDP representation ${\mathbf{P}}_{tar}$ of the target pose and a character sheet $\mathbf{S}_{ref}$, as the inputs. When generating more than one pose, we feed different UDPs and use the same reference character sheet.

The input UDP representation can be produced by a UDP Detector from reference images. For interactive applications like games, the existing physics engine can be used as a drop-in replacement for the UDP Detector to compute body and cloth dynamics for the anime character directly.

\noindent\textbf{Renderer.} We apply the following modifications to the U-Net~\cite{ronneberger_u-net_2015}:

\noindent1)~As shown in Figure \ref{fig:acs}(c), we rescale and concatenate UDP to each block of the decoder. This input strategy aims to reuse the encoder's results and further allow highly efficient inference when there are multiple target UDPs when generating an animation video of a character.

\noindent2)~Due to the spatial misalignment of the local and the remote branches, we use flow fields to align the features.
Specifically, we approximate a flow field $\textbf{f}$ as two channels and warp the features of other channels according to the estimated flow~\cite{zhou2016view,dvf,hu2023dmvfn}. This operator enhances the long-range look-up ability for CINN. 

\noindent3)~We use the CINN method in the decoders of the network. We split the original up-sampling output feature channels by half, one for the remote branch and the other for the local branch. 
Firstly, we warp the features of the remote branches to align with the local features. Then the output features from all remote branches are averaged to be concatenated with the encoder output. The concatenated feature will be fed into the next block~(illustrated in \textbf{Appendix}). Formally, to aggregate local feature $Feat_l$ and remote feature $Feat_r^i(0<i<k)$:
\begin{equation}
\textbf{f}_r^i = Conv_{3\times 3}(PReLU(Conv_{3\times 3}(Feat_r^i))),
\end{equation}
\begin{equation}
Feat_l^{*} = Feat_l + \sum_{i=0}^{k-1}\backwardwarp(Feat_r^i, \textbf{f}_r^i) / k, 
\end{equation}
where we denote the pixel-wise backward warping~(remapping) as $\backwardwarp$. $PReLU$ and $Conv_{3\times 3}$ represent PReLU activation function~\cite{he2015delving} and $3\times 3$ convolution. 

The last decoder block will collect averaged output features from all previous decoder blocks in all branches and output the final generated image $\widehat{I}_{tar}$.  

\noindent\textbf{UDP Detector.} 
While training with the synthesized dataset, UDP $\widehat{\mathbf{P}}_{tar}$ can be directly obtained. As for the hand-drawn dataset, we design a UDP Detector to estimate UDPs $\widehat{\mathbf{P}}_{tar}$ from hand-drawn images.
The UDP Detector is a simple U-Net~\cite{ronneberger_u-net_2015} consists of a ResNet-50~(R50)~\cite{resnet} encoder and a decoder with $5$ residual blocks. It is trained jointly with the renderer in an end-to-end manner. We share the weight of the renderer's encoder and the UDP Detector's encoder.
\section{Experiments}
\subsection{Training Strategy}

We train CoNR with $m$ sub-networks~(views) on our dataset. 
To create one training sample, we randomly select a character and then randomly select $m+1$ arbitrary poses. These images are split as $m$ image inputs ~(character sheet) ${I}_1 \cdots {I}_m\in {S}_{ref}$ and one image of the target pose as the ground truth of CoNR's final output. 

We paste them onto $k$ random backgrounds and feed them into the UDP Detector. We use the average of the $k$ UDP detection results $\widehat{{P}}_{i}$ of the same target pose, $ \widehat{{P}}_{tar} = 1/k \sum_{i=1}^{k} \widehat{{P}}_{i}$, as the final UDP detection results. We compute losses at both the output of the detector and the end of the CoNR pipeline. We use L1 loss on the landmark encodings and binary cross-entropy~(BCE) loss on the mask to train the detector if the ground truth UDP ${P}_{tar}^{GT}$ is available:
\begin{equation}
	\mathcal{L}_{udp} = || \widehat{{P}}_{tar} - {P}_{tar}^{GT} ||_1, 
\end{equation}

\begin{equation}
	\mathcal{L}_{mask} = BCE ( sgn(\widehat{{P}}_{tar}), sgn({P}_{tar}^{GT})),
\end{equation}
\noindent where $sgn$ indicates that a point is from the body surface or background.
We use a consistency loss $\mathcal{L}_{cons}$ by computing the standard deviation of $k$ UDP Detector outputs: 
\begin{equation}
	\mathcal{L}_{cons} = \sqrt{\frac{1}{k-1}\sum_{i=1}^k(\widehat{{P}}_{i}-\widehat{{P}}_{tar})^2}.
\end{equation}
At the end of the collaborated renderer, we use L1 loss and feature loss~\cite{ledig2017photo} which is based on a pre-trained 19-layer VGG~\cite{vgg} network to supervise the reconstruction in the desired target pose, denoted as $L_{photo}$ and $L_{vgg}$.

\noindent The UDP Detector and renderer are trained end-to-end simultaneously. The total loss function is the sum of all losses:
\begin{equation}
    \mathcal{L} = \mathcal{L}_{udp}+\mathcal{L}_{cons}+\mathcal{L}_{mask}+ \mathcal{L}_{photo}+\alpha \mathcal{L}_{vgg},
\end{equation}
\noindent where the hyper-parameter $\alpha$ of feature loss is from previous work~\cite{zhang2018single}.

Our model is optimized by AdamW~\cite{loshchilov_decoupled_2019} with weight decay $10^{-4}$ for $224K$ iterations. We choose $m=4, k=4$ during training unless otherwise specified. The training process uses a batch size of $24$ with all input resolutions set to $256\times256$. Model training takes about three weeks on four GPUs. 

\subsection{Result}

For quantitative evaluation on the validation set, we measure $L_{udp}$ and $L_{photo}$ which is the averaged L1 distance between predictions and ground truth. We further use LPIPS~\cite{lpips} to measure the perceptual quality.

% \subsection{Comparative Study}
\label{sec:model_ablation}

\begin{table}[tb]
\centering
\begin{tabular}{ccccccc}
\hline
\multirow{2}{*}{Setting} & \multicolumn{2}{c}{$112K$ iter} & \multicolumn{2}{c}{$224K$ iter} \\
                         & $\mathcal{L}_{photo}$          & LPIPS         & $\mathcal{L}_{photo}$           & LPIPS  \\ \hline
						 $m=1, n=1$~~                    & 0.0247          & 0.0832            & 0.0238         & 0.0801         \\
						 $m=4, n=1$~~  & 0.0249 & 0.0865 &	0.0237 &	0.0827 \\
						 $m=1, n=4$~~                      & 0.0219          & 0.0798           & 0.0211     & 0.0764     \\
						 $m=4, n=4$~~                & \textbf{0.0187}          & \textbf{0.0659}            & \textbf{0.0179}           & \textbf{0.0612}           \\
%						$m=4, n=8$  &	0.0185  &	0.0631  &	0.0172  &	0.0582\\
						 
\hline
\end{tabular}
\caption{
		\textbf{Comparison on the number of input reference images. }We use character sheets of $m$ reference images to train the CoNR model, and then use character sheets of $n$ reference images to evaluate the trained model.
}
\label{tab:viewcount}
\end{table}

\begin{figure}[tb]
\centering
\includegraphics[width=0.9\linewidth]{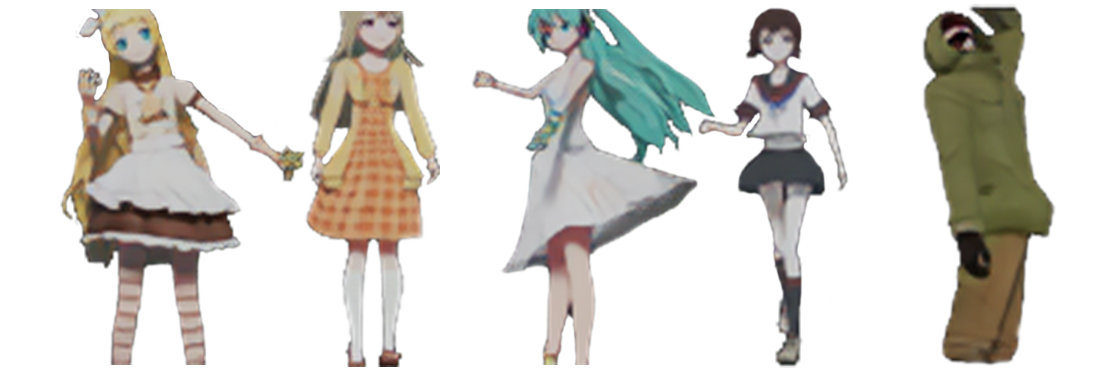}
\includegraphics[width=0.75\linewidth]{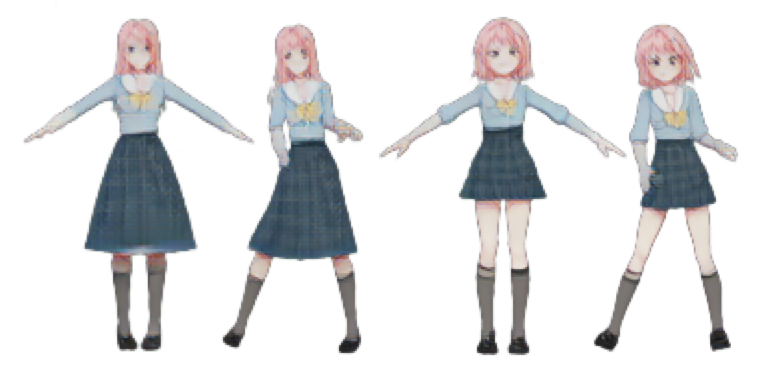}
\caption{\textbf{First row}: Inference results on validation dataset. \textbf{Second row}: Inference results with the same character sheet input ${S}_{ref}$ on different body structure ${P}_{tar}$.}
\label{fig:teaser}
\end{figure}

\noindent\textbf{Inference Visual Effects.} CoNR can cope with the diverse styles of anime, including differences between synthesized and hand-drawn images. Figure~\ref{fig:teaser} lists
some random CoNR outputs on the validation split. With the same character sheet input ${S}_{ref}$, we replaced the provided UDP ${P}_{tar}$, and the results of CoNR changed accordingly. Notably, CoNR goes beyond a naïve correspondence by absolute UDP value~(transfer between different cloths).

\begin{figure}[t]
\begin{minipage}[t]{0.38\linewidth}
  \centering
  \includegraphics[width=1\linewidth]{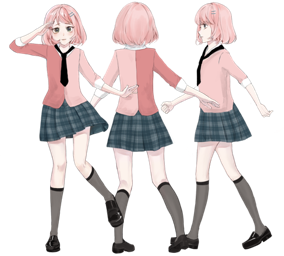}
  \centerline{Character Sheet}
  \end{minipage}
 \begin{minipage}[t]{0.35\linewidth}
  \centering
  \includegraphics[width=1\linewidth]{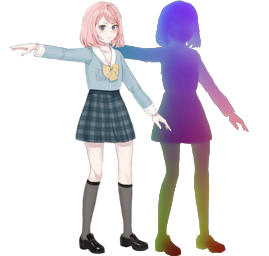}
  \centerline{UDP}
  \end{minipage}
  \begin{minipage}[t]{0.235\linewidth}
  \centering
  \includegraphics[width=1\linewidth]{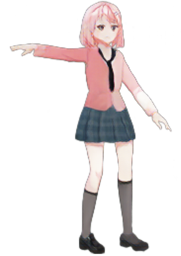}
  \centerline{Generated}
  \end{minipage}
\caption{\textbf{Transfer appearance based on the character sheet.} This UDP comes from UDP Detector rather than 3D softwares.}
\label{fig:transfer}
\end{figure}

CoNR with UDP Detector can also be used to achieve similar results to style transfer methods when running a textured image at the reference pose. As shown in Figure~\ref{fig:transfer}, UDP detection of one character can be used to render another character. The CoNR inference pipeline for the anime (or game) production will usually be without the UDP Detector.

\noindent\textbf{Effectiveness of the Collaboration.} CoNR uses a dynamically-sized set of reference inputs during training and inference. Table~\ref{tab:viewcount} shows that using additional views ($m>1$) during training will enhance the quality of generated images. 
On the opposite, removing images from the character sheet will reduce the coverage of the body surface. When CoNR is trained with $m=1$, CoNR can not provide a reasonable solution. For example, given the character's backside, it is hard to imagine the frontal side. In this case, the photo-metric reconstruction and feature losses may encourage the network to learn a wrong solution. Therefore even if enough information in the $n>1$ images is provided during inference, the network may not generate the target image accurately.
Similarly, keeping $m=4$ while reducing the number of inference views~($n=1$) will also harm the quality of CoNR. 

\begin{figure*}[tb]
	\centering
	\includegraphics[width=18cm]{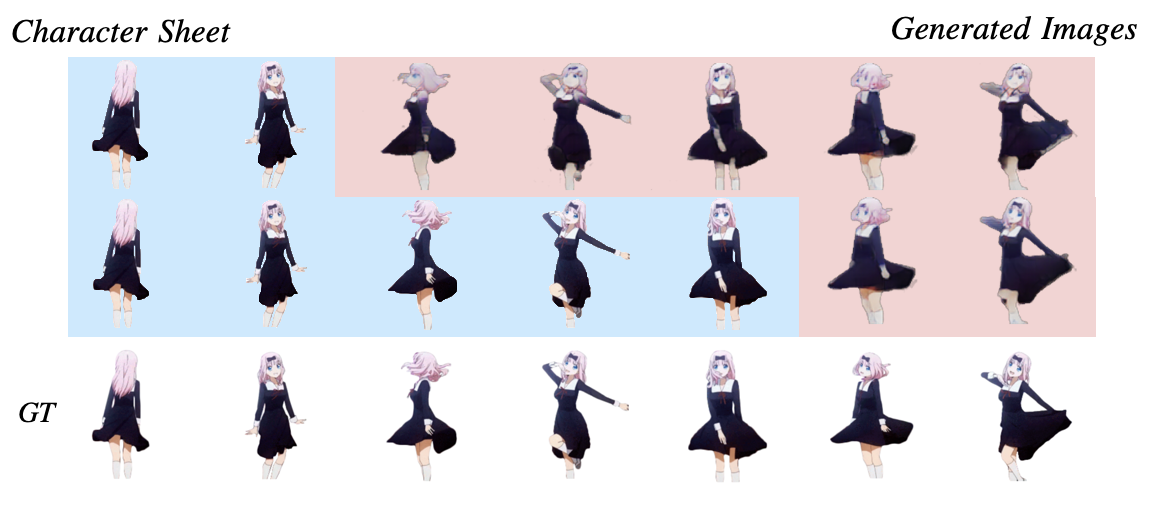}
	\caption{\textbf{Effectiveness of the collaboration.} We perform a reconstruction experiment with Chika Dance, which is a high-quality rotoscoping animation (in which body and clothing motions are drawn according to real characters) ensuring that the ground truth is reasonable. The last row shows $8$ ground truth frames ${I}^{GT}_{i}\in {S}^{GT}_{vid}$ from a video. In this experiment, the UDPs are estimated from ground truth frames by a trained UDP Detector. The first two rows show the input and output images of CoNR. The used subsets of character sheet ${S}_{ref} \subset {S}^{GT}_{vid} $ are marked using the blue background. Generated images for novel poses are marked using the red backgrounds.  }\label{fig:abl_inferviewcount}
\end{figure*}

As shown in Figure \ref{fig:abl_inferviewcount}, CoNR can leverage information distributed across all images to produce better visual effects.
This allows CoNR to scale from image synthesis, when only a few shots of the character are given, to image interpolating when a lot of shots are available. 
The example in Figure~\ref{fig:abl_inferviewcount} shows the behavior of CoNR when it does not have enough information to draw missing parts correctly. Even if very few reference images are provided, the target pose that is similar to some reference will be accurate. Furthermore, users may iterate on the results and feed them back into CoNR to accelerate anime production.

\subsection{Comparison with Related Work} 
As CoNR provides a baseline for a new task, we admit that direct comparison with related works can either be impossible or unfair. We still try to include comparisons, only to show how our task is related to other existing tasks.

\begin{figure}[tb]
	\centering
	\includegraphics[width=8.4cm]{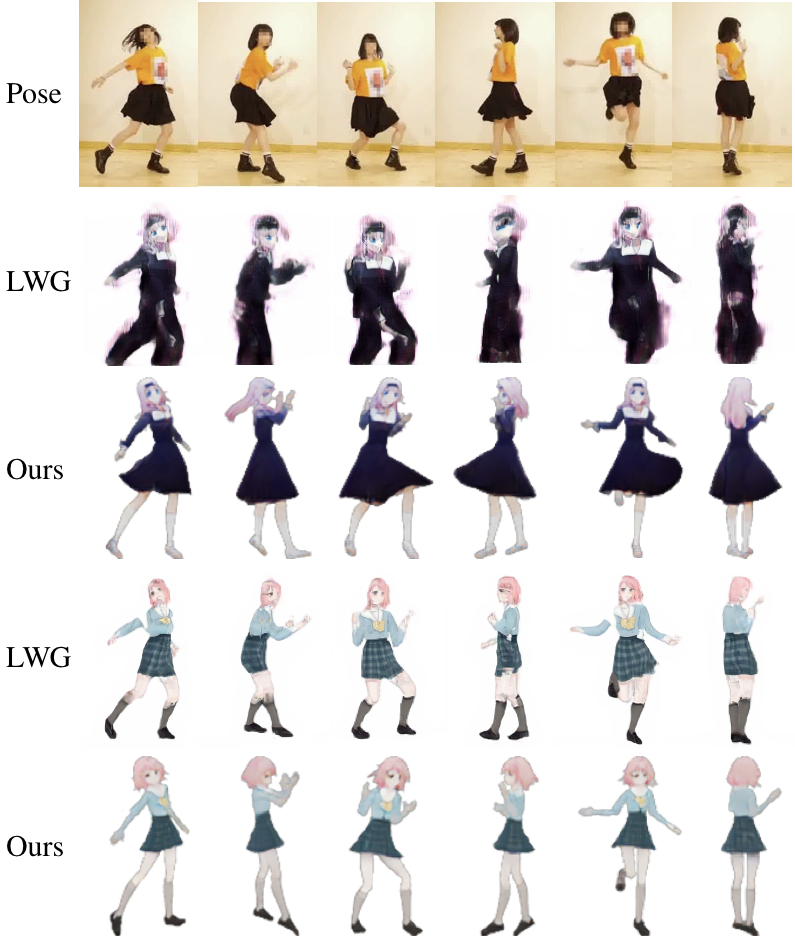}
	\caption{\textbf{Conceptual comparison with pure image-based method.} We compare the results of CoNR with the results of Liquid Warping GAN (LWG)~\protect\cite{liu_liquid_nodate} when trying to resemble the target poses. We use 3D model editor software to generate pose-related UDPs based on the real person then make CoNR render the texture. And LWG uses the real person images as input.}\label{fig:compare_smpl}
\end{figure}

\begin{figure}
	\centering
	\includegraphics[width=8.4cm]{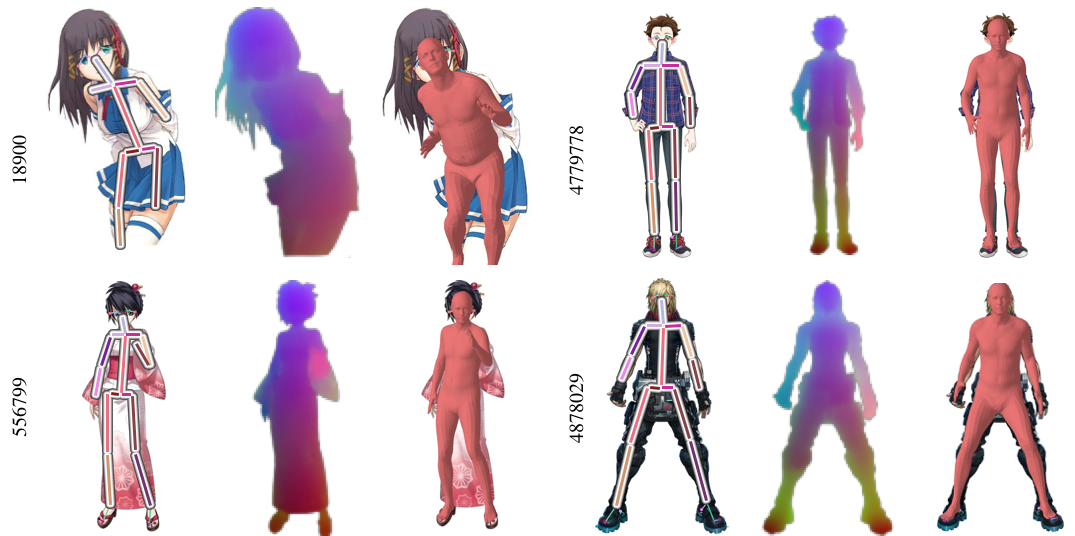}
	\caption{\textbf{Comparision between detection results of hand-drawn anime character images using OpenPose~\protect\cite{paf}, the UDP Detector and SMPLify~\protect\cite{bogo2016keep}.} The images are from the validation split of the hand-drawn dataset~\protect\cite{danbooru2020} with the ID number. The detected SMPL body can not fully handle the diverse body shapes of anime characters. UDP Detector produces relatively reasonable results.}\label{fig:conr_suppl}
\end{figure}

\noindent\textbf{Human Pose Synthesis.} We compare the results produced by CoNR to a real-world digital human system~\cite{liu_liquid_nodate} using the same target poses as used in their demo. We use two character sheets~\footnote{One anime character is from \url{www.youtube.com/watch?v=m6k_t8yEyvE} and another character is illustrated by an amateur artist.} with both pose and appearance unseen during training. One contains the same $4$ images shown in Figure~\ref{fig:acs}, the other character sheet taken from a random video from the internet as used in Figure~\ref{fig:abl_inferviewcount}.
Figure~\ref{fig:compare_smpl} indicates that the long skirt prevents a high accuracy estimation of the leg joints and that parametric 3D human models like SMPL may not handle the body shape of anime characters correctly. Further diagnosis shows that parametric 3D human models may not handle anime's diverse clothing and body shape, as shown in Figure~\ref{tab:comparison_detector}. CoNR can produce images at desired target poses with better quality. 

Image synthesis pipelines starting with human pose representations are theoretically inapplicable on anime data, as the character's diverse body structure, clothing, and accessories cannot be reasonably represented. Using human pose synthesis methods in anime, we may lose artistic control over garments and fine details, which is crucial in anime creation workflows.

\begin{table}[tb]
\centering
\begin{tabular}{ccccc}
\hline
\multirow{2}{*}{Setting} & \multicolumn{2}{c}{$112K$ iter} & \multicolumn{2}{c}{$224K$ iter} \\
            \cmidrule(lr){2-3}\cmidrule(lr){4-5}             & $\mathcal{L}_{udp}$           & $\mathcal{L}_{mask}$          & $\mathcal{L}_{udp}$            & $\mathcal{L}_{mask}$            \\ \hline
Original U-Net  & 0.1247          & 0.0856            & Fail~            & Fail~            \\
U-Net+R34                      & 0.1051          & 0.1068           & 0.1004     & 0.0747    \\
\textbf{U-Net+R50~~}                    & \textbf{0.0969}          & \textbf{0.0792}            & \textbf{0.0971}           & \textbf{0.0736}           \\
\hline
\end{tabular}
\caption{\textbf{Ablation on UDP Detector with different backbones}.}
\label{tab:comparison_detector}
\end{table}

\begin{table}[tb]
\centering
\begin{tabular}{ccccccc}
\hline
\multirow{2}{*}{U-Net~} & \multirow{2}{*}{Warping~} & \multirow{2}{*}{CINN~} & \multirow{2}{*}{R50~} & \multicolumn{2}{c}{$224K$ iter} \\
                 \cmidrule(lr){5-6}       & & & & $\mathcal{L}_{photo}$          & LPIPS             \\ \hline
$\checkmark$ & & &             & 0.0311           & 0.1038           \\
$\checkmark$ & $\checkmark$ &             & &  0.0308           & 0.1036           \\ 
$\checkmark$ & & $\checkmark$ & $\checkmark$      & 0.0286           & 0.0977           \\$\checkmark$ & $\checkmark$ & $\checkmark$ &                              & 0.0180  & 0.0612  \\
$\checkmark$ & $\checkmark$ & $\checkmark$ & $\checkmark$       & \textbf{0.0179}  & \textbf{0.0612}           \\ \hline
\end{tabular}
\caption{
\textbf{Ablation on Renderer.} The warping operation is performed among all branches.
}
\normalsize
\label{tab:comparison_renderer}
\end{table}

\noindent\textbf{Style Transfer.} Style transfer typically refers to applying a learned style from a certain domain~(anime images) to the input image taken from the other domain~(real-world images)~\cite{chen2019animegan}. The models usually treat the target domain as a kind of style and require extensive training to remember the style.
Some methods use a single image to provide a style hint during the inference. For example, one could use Swapping Auto Encoders~\cite{park2020swapping}, a recent style-transfer method to swap the textures or body structures between two characters.
Although the model has a lot of parameters (3$\times$ the size of CoNR), our comparison shows that it is still not enough to remember and reproduce the diverse sub-styles of the textures, pose, and body structure in the domain of anime.

\subsection{Ablation Study}

\begin{table}[tb]
\centering
\hspace{-55pt}
\begin{minipage}[t]{0.130\linewidth}
	\centering
    \includegraphics[width=0.6\linewidth]{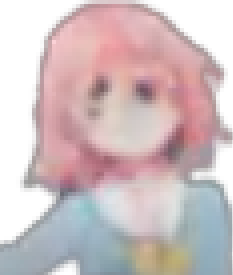}
    
    1 time
    \includegraphics[width=0.6\linewidth]{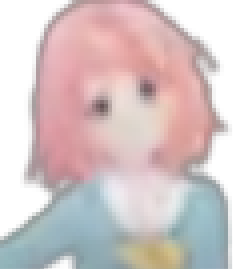}
    
    3 times
\end{minipage}
\begin{minipage}[t]{0.600\linewidth}
\vspace{-5mm}
\hspace{7pt}
\setlength{\tabcolsep}{1.6mm}
{
\vspace{-1mm}
\begin{tabular}{ccccccc}
\hline
\multirow{2}{*}{Messaging} & \multicolumn{2}{c}{112K iter} & \multicolumn{2}{c}{224K iter} \\
            \cmidrule(lr){2-3}\cmidrule(lr){4-5} 
                         & $\mathcal{L}_{photo}$          &  LPIPS          & $\mathcal{L}_{photo}$          &  LPIPS             \\ \hline
						 -   & 0.028          & 0.107            & 0.026         & 0.099         \\
						 1 time   & 0.019 & 0.066 & 0.018 & 0.063 \\
						 3 times   & \textbf{0.018}  & \textbf{0.065} & \textbf{0.017}     & \textbf{0.061}     \\
					
\hline
\end{tabular}
}
\end{minipage}
\caption{\textbf{Ablation on message passing}.}
\label{tab:mess}
\end{table}

We perform ablation studies on the UDP Detector, renderer, loss functions and message passing. Table~\ref{tab:comparison_detector} shows that UDP representation can be inferred from images using a U-Net~\cite{ronneberger_u-net_2015}. An original U-Net, which takes a concatenated tensor of $4$ reference images and the target UDP as the input, does not provide acceptable results on this task, as shown in Table~\ref{tab:comparison_renderer}. The proposed CoNR method with both the feature warping and the CINN method significantly increases the performance, thus establishing a stronger baseline for the proposed task. We further ablate the loss functions in Table~\ref{tab:loss}. In Table~\ref{tab:mess}, we perform ablations on the number of message passing for the CoNR. In default, the sub-networks communicate at three different feature levels. More ghosting can be observed when sub-networks communicate less than three times. 

\begin{table}[tb]
\centering

\begin{tabular}{ccccccccc}
\hline
\multirow{2}{*}{Setting} & \multicolumn{4}{c}{$224K$ iter} \\
               \cmidrule(lr){2-5} & $\mathcal{L}_{udp}$            & $\mathcal{L}_{mask}$ & $\mathcal{L}_{photo}$           & LPIPS             \\ \hline
w/o $\mathcal{L}_{mask}$  & 0.162          & 0.880   &   - & -       \\
w/o $\mathcal{L}_{photo}$  & -          & -   &   0.023~ & 0.084~~     \\
w/o $\mathcal{L}_{vgg}$  & -          & -   &   0.028~ & 0.158~~      \\
\textbf{CoNR}                    & \textbf{0.097}          & \textbf{0.079} & \textbf{0.019} & \textbf{0.066}~~   \\
\hline
\end{tabular}
\caption{\textbf{Ablation on loss functions.}}
\label{tab:loss}
\end{table}

% \mathcal{L}_{udp}+\alpha \mathcal{L}_{mask}+\beta LPIPS+\gamma \mathcal{L}_{photo}+\theta \mathcal{L}_{cons}
\section{Limitations}
\label{sec:limitation}
% CoNR is unable to model the environmental lighting effects. 
The inputs of CoNR can not provide any information about the environmental or contextual information that could be utilized to infer lighting effects. Users may have to look for sketch relighting techniques ~\cite{zheng_learning_2020}. The generation results of finer layering and structuring are also to be studied.

CoNR is unable to model the dynamics of the character. 
The CoNR model accepts target poses detected from a video of a character with a body shape similar to ${S}_{ref}$, and could inherit the dynamics. However, it requires such a pose sequence beforehand. 
To bypass the UDP Detector, we can rely on additional technologies like garment captures~\cite{bradleyMarkerlessGarmentCapture2008}, physics simulations~\cite{baraffLargeStepsCloth1998}, learning-based methods~\cite{tiwariNeuralGIFNeuralGeneralized2021} or existing 3D animation workflows to obtain a synthesized UDP ${P}_{tar}$.
CoNR focuses on the rendering task. Obtaining a suitable 3D mesh with all the body parts rigged and all clothing computed with proper dynamics is beyond the scope of this paper. Using ${P}_{tar}$ from an inappropriate body shape may lead to incorrect CoNR results.

The dataset may not fully follow the distribution of anime characters in the wild.
The collected dataset contains only human-like anime characters from 2014 to 2018. 
As the character meshes are aligned using joints, the models trained on this dataset may not be applied to animal-like characters. Research on broader datasets will be the future work. 

\section{Conclusion}
In this paper, we explore a new task to render anime character images with the desired pose from multiple images in character sheets. We develop a UV-mapping-free method, CoNR, achieving encouraging effectiveness. We show the potential of this data-driven method in assisting animation art. We hope the method and the datasets presented in this paper will inspire our community and further researchers.

{\small
\bibliographystyle{named}
\bibliography{mybib}
}
\section{Appendix}
\subsection{Postscript}
Zuzeng Lin explored the idea of assisting anime creation with neural networks. He proposed CoNR as a baseline to solve consistency and artistic control issues at the end of the 2020. He did most of the experiments and wrote the first draft. He quit subsequent submissions because of receiving the negative reviews from CVPR2022 and ECCV2022. He was therefore no longer involved in the submissions to AAAI2023 and IJCAI2023, which were made by other authors. They revised the paper to its present state with various demos. Zuzeng appreciates the discussions with many people who are interested in this project and Live3D public beta users in Sept. 2021.

\subsection{Dataset Details}

The synthesized part of the dataset contains multiple-pose and multiple-view images. Each of them is paired with UDP annotations.
A randomly-sampled subset of 3D mesh data obtained in the same way as in  ~\cite{khungurn_pose_2016} was used in this work, which contains $2,000$ anime characters with more than $200$ poses. The scope of permissions was confirmed before using existing assets in this work. The dataset contribution in this work is mostly about years of manual data clean-up, which includes converting mesh formats, restructuring the bone hierarchy, and aligning the A-poses before synthesizing RGB images for each pose, and their corresponding UDPs with MikuMikuDance~(MMD) software.

The final dataset also contains unlabeled hand-drawn samples and one-shot samples, \emph{i.e.} characters with the only single available pose, which are randomly sampled from existing datasets~\cite{danbooru2020,pixiv}. We keep the licenses of all 3D models and motions meet the usage and distribution requirements. 

\subsection{Experiment Details}

We make full use of the dataset by performing semi-supervised learning. For the UDP Detector, we skip the $\mathcal{L}_{udp}$ when we do not have ground truth of the UDP. We will keep the $\mathcal{L}_{
cons}$ to encourage the UDP Detector to make a same prediction of one character under $k$ random augmentations.

As for the renderer, we have to reuse the image of the same pose multiple times to fill all the $m$ views if the provided images in the character sheet are not enough. Random-crop augmentation is applied to $ \mathbf{I}_{m} \in \mathbf{S}_{ref}$,  $\widehat{\mathbf{P}}_{tar} $ and $\mathbf{I}_{tar}^{gt}$. With the random-crop augmentation, in order to generate the image of body parts in the right position, CoNR has to learn a cross-modality feature matching instead of a trivial solution by simply selecting one of the provided images as the output. 

\subsection{Comparison with Style Transfer}
\begin{figure*}
	\centering
	\includegraphics[width=16cm]{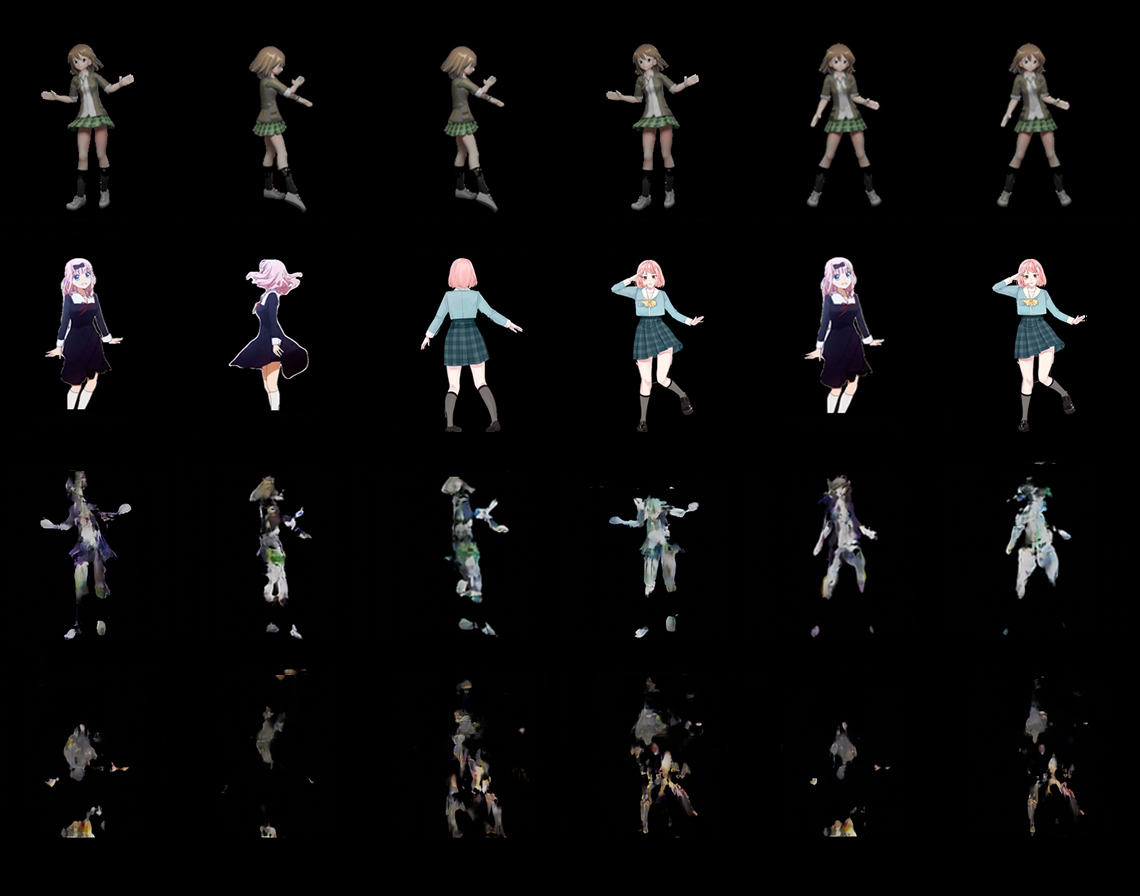}
	\caption{\textbf{Evaluation results of Swapping Autoencoder (SwapAE)~\protect\cite{park2020swapping} for Deep Image Manipulation.} We trained a PyTorch implementation of SwapAE~\protect\cite{park2020swapping} with pairs of images of different characters in our dataset. This figure, from the first to the last row, shows (a) the target pose image, (b) the reference image, (c) pose of \textbf{a} and texture of \textbf{b}, fused by SwapAE, (d) pose of \textbf{b} and texture of \textbf{a}, fused by SwapAE. }\label{fig:swapping}
\end{figure*}

To best of our knowledge, fusion of pose and texture for virtual character cannot be solved directly by implicit style transfer. As shown in Figure~\ref{fig:swapping}, SwapAE~\cite{park2020swapping} fails in this scene.

\subsection{Relationship with Neural Digital Human System}

Some neural rendering systems can produce demos which may be similar to CoNR visually. For example, ANR~\cite{raj2021anr} uses UV mapping, which requires the assumption that the characters essentially have a similar topology. The main difference is that CoNR may deal with more general situations such as hair and long skirts. In addition, real-world video is constrained by geometric consistency, but painting creation is not. For 2D animation, the existing methods can not even accurately identify the skeleton. This domain gap is a part of our motivation to explore UDP tailored for animation.
\end{document}